\documentclass[runningheads]{llncs}
\usepackage{comment} 
\usepackage{graphicx}
\usepackage{soul} 
\usepackage{siunitx}
\usepackage{graphicx}
\usepackage{caption}
\usepackage{booktabs, makecell}
\usepackage{blindtext}
\usepackage{makecell}
\usepackage{float}
\usepackage[caption=false]{subfig}
\usepackage{lipsum} 
\usepackage{changepage}
\usepackage{amsmath}
\usepackage{xcolor}

\newcommand{\lia}[1]{{\color{black}#1}}
\newcommand{\francesco}[1]{{\color{black}#1}}
\begin{document}
\title{Slicing and dicing soccer: automatic detection of complex events from spatio-temporal data}
\titlerunning{Slicing and dicing soccer}
%
\author{Lia Morra\inst{1}\orcidID{0000-0003-2122-7178} \and
Francesco Manigrasso\inst{1} \and
Giuseppe Canto\inst{1} \and
Claudio Gianfrate\inst{1} \and
Enrico Guarino\inst{1} \and
Fabrizio Lamberti\inst{1}\orcidID{0000-0001-7703-1372}}
\authorrunning{L. Morra et al.}
%
\institute{Politecnico di Torino, Department of Control and Computer Engineering, Turin IT 10129, Italy
\email{\{lia.morra\}@polito.it}\\}
\maketitle              
\begin{abstract}
The automatic detection of events in sport videos has important applications for data analytics, as well as for broadcasting and media companies. This paper presents a comprehensive approach for detecting a wide range of complex events in soccer videos starting from positional data. The event detector is designed as a two-tier system that detects \textit{atomic}  and \textit{complex events}. Atomic events are detected based on temporal and logical combinations of the detected objects, their relative distances, as well as spatio-temporal features such as velocity and acceleration. Complex events are defined as temporal and logical combinations of atomic and complex events, and are expressed by means of a declarative Interval Temporal Logic (ITL).  The effectiveness of the proposed approach is demonstrated over 16 different events, including complex situations such as tackles and filtering passes. By formalizing events based on principled ITL, it is possible to easily perform reasoning tasks, such as understanding which passes or crosses result in a goal being scored. To counterbalance the lack of suitable, annotated public datasets, we built on an open source soccer simulation engine to release the synthetic SoccER (Soccer Event Recognition) dataset, which includes complete positional data and annotations for  more than 1.6 million atomic events and 9,000 complex events. \lia{The dataset and code are available at https://gitlab.com/grains2/slicing-and-dicing-soccer.}

\keywords{Sport analysis \and Event Detection \and Interval Temporal Logic \and Computer graphics}
\end{abstract}
\section{Introduction}

Data-driven sport video analytics attracts considerable attention from academia and industry. This interest stems from the massive commercial appeal of sports programs, along with the increasing role played by data-driven decisions in soccer and many other sports \cite{shih2017survey}. We focus here on the challenging problem of temporal event recognition and localization in soccer, which requires considering the positions and actions of several players at once.

\lia{Sports analytics systems relies on a variety of data sources for event detection, including broadcast videos \cite{rematas2018soccer,khan2018soccer,shih2017survey}, multi-view camera setup \cite{shih2017survey,Pettersen2014} and wearable trackers and sensors \cite{richly2016recognizing,cannavo2019}. Large outdoor soccer stadiums are usually equipped with multiple wide-angle, fixed position, synchronized cameras. This setup is particularly apt at event recognition as the spatio-temporal location of all players can be inferred in an unobtrusive and accurate fashion, without resorting to ad-hoc sensors, as will be detailed in Section \ref{sec:archi}. } 

Previous attempts at sports event recognition \lia{fall in two main categories}: machine learning techniques applied to spatio-temporal positional data \cite{richly2016recognizing,networks,cannavo2019} or knowledge-based systems based, e.g., on finite state machines, fuzzy logic or first-order logic \cite{shih2017survey,khan2018soccer}. The latter approach has several advantages in this context: it does not require large training set, takes full advantage of readily available domain knowledge, and can be easily extended with reasoning engines.

We propose here a comprehensive event detection system based on Interval Temporal Logics (ITL). \lia{Khan et al.  applied a similar approach to  identify events of interest in broadcast videos \cite{khan2018soccer}: the distance-based event detection system takes as input bounding boxes associated with a confidence score for each object category, and applies first-order logic to identify simple and complex events. Complex events combine two or more simple events using logical (AND, OR) or temporal (THEN) operators. }

\lia{Our work extends previous attempts in literature \cite{khan2018soccer} in several ways. First, we work on spatio-temporal data instead of broadcast videos: we are thus able to detect events that require the position of multiple players at once (e.g. filtering pass), or their location within the field (e.g., cross). We thus cover a much wider range of events, determining which can be accurately detected from positional data, and which would need integration with other visual inputs (e.g., pose estimation). Lastly, we extend existing rule-based system by using more expressive ITLs, which associate to each event a time interval and are capable of both qualitative and quantitative ordering.}

\lia{A severe limitation for developing sports analytics systems is} the paucity of available datasets, which are usually small and lack fine-grained event annotations. This is especially true for multi-view, fixed setups comparable to those available in modern outdoor soccer stadiums \cite{Pettersen2014}. A  large scale dataset was recently published based on broadcast videos \cite{giancola2018soccernet}, but annotations include only a limited set of events (Goal, Yellow/Red Card, and Substitution).  

With the aim of fostering research in this field, we have generated and released the  synthetic Soccer Event Recognition (SoccER) dataset, based on the open source Gameplay Football engine. The Gameplay Football engine was recently proposed as a training gym for reinforcement learning algorithms \cite{kurach2019google}. We believe that event recognition can similarly benefit from this approach, especially to explore aspects such as the role of reasoning and the efficient modeling of spatial and temporal relationships.
We used the dataset to demonstrate the feasibility of
\lia{our approach}, achieving precision and recall higher than 80\% on most events.

The rest of the paper is organized as follows. Section \ref{sec:dataset} introduces the SoccER dataset. In Section \ref{sec:Eve}, the event detector is described. Experimental results are presented in Section \ref{sec:results} and discussed in Section \ref{sec:discussion}.


\section{The SoccER Dataset}
\label{sec:dataset}
\subsection{Modified Gameplay Football engine}
We designed a solution to generate synthetic datasets starting from the open source Gameplay Football game \cite{Gameplay_Football}, which simulates a complete soccer game, including all the most common events such as goals, fouls, corners, penalty kicks, etc. \cite{kurach2019google}.  While the graphics is not as photorealistic as that of commercial products, the game physics is reasonably accurate and, being the engine open source, it can be inspected, improved and modified as needed for research purposes. The opponent team is controlled by means of a rule-based bot, provided in the original Gameplay Football simulator \cite{kurach2019google}.

\begin{figure}[tb]
	\centering
		\includegraphics[width=7cm]{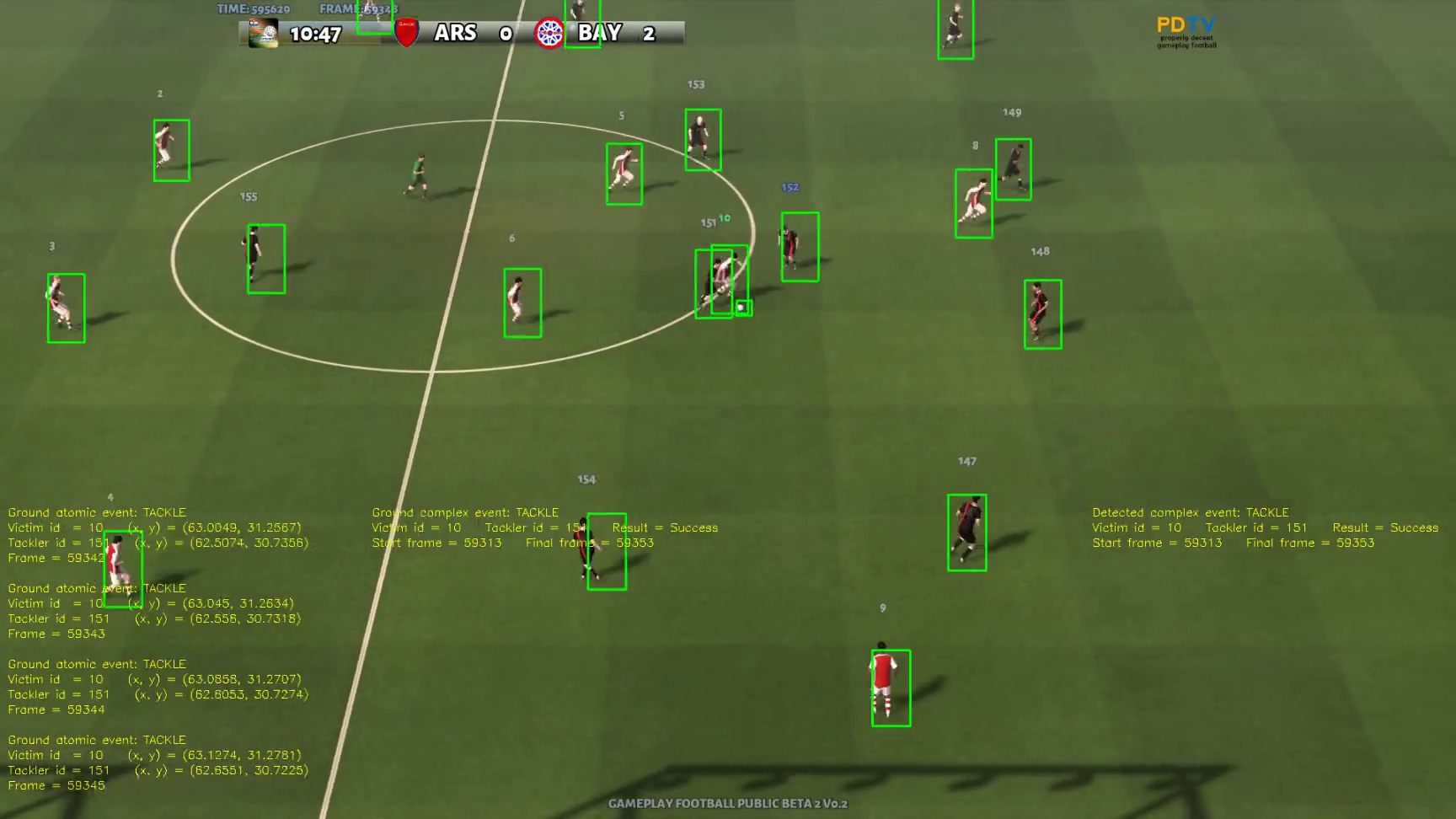}
		\caption{Example of scene generated by the Gameplay Football engine, with superimposed ground truth bounding boxes and IDs of each player and the ball. The ground truth and detected events are also overlaid on the bottom of the scene: in this frame, a tackle attempt is correctly detected.}
		\label{Figure:game_frame}
	\end{figure}

For each time frame, we extract the positions and bounding boxes of all distinct 22 players and the ball, the ground truth event annotation and the corresponding video screenshots. We adopt the same field coordinate system used in the  Alfheim dataset, which includes the position of players obtained from wearable trackers \cite{Pettersen2014}. All the generated videos have a resolution of 1920$\times$1080 pixels (Full HD) and frame rate of 30 fps. An example of generated frame is reported in Fig. \ref{Figure:game_frame}. We envision that event detectors can be trained and tested directly on the generated positional data, focusing on the high-level relational reasoning aspects of the soccer game, independently of the performance of the player detection and tracking stage \cite{rematas2018soccer,khan2018soccer}.

\subsection{Events and generated datasets}

Events are automatically logged by the game engine in order to generate the ground truth annotation. We define the notion of event based on previous work by Tovinkere et al. \cite{SoccerDetectionEvent} and Khan et al. \cite{khan2018soccer}. 
Similarly to \cite{khan2018soccer}, we distinguish between \textit{atomic} and \textit{complex} events, with a slightly different approach (as discussed in the next sub-section). Atomic events are those that are spatio-temporally localized, whereas complex (compound) events are those that occur across an extended portion of the field, involve several players or can be constructed by a combination of other events. Stemming from this difference, an atomic event is associated to a given time frame, whereas a complex event is associated to a time interval, i.e., to a starting and ending frame. Atomic events include ball possession, kicking the ball, ball deflection, tackle, ball out, goal, foul and penalty. Complex events include ball possession, tackle, pass and its special cases (filtering pass, cross), shot and saved shot. A complex ball possession, or tackle, event corresponds to a sequence of consecutive atomic events that involve the same players. The ground truth also includes examples of chains of events, such as a pass, filtering pass or cross that led to a goal.  

The annotations are generated leveraging information from the game engine bot, independently from the detection system: different finite state machines detect the occurrence of several types of events based on the decisions of the bot or the player, their outcomes and the positions of all the players. The definition of each event was double-checked against the official rules of the Union of European Football Association (UEFA), and the annotations were visually verified. 

For the present work, eight matches were synthesized through various modalities (player vs. player, player vs. AI, AI vs. AI), for a total of 500 minutes of play with 1,678,304 atomic events and 9,130 complex events, divided in a training and testing set as reported in Table~\ref{table:list_of_events_dataset}.  
\lia{The game engine and dataset are available at https://gitlab.com/grains2/slicing-and-dicing-soccer.}

\begin{table*}[t] 
\begin{center}
\begin{tabular}{|l|c|c|}
\hline
\textbf{Atomic event} & \textbf{Train Set}  & \textbf{Test set}  \\
\hline
\emph{KickingTheBall} & 3,786 &  3,295 \\
\hline
\emph{BallPossession} & 812,086 & 797,224 \\
\hline
\emph{Tackle} & 34,929& 26,286  \\
\hline
\emph{BallDeflection} & 172 & 78 \\
\hline
\emph{BallOut} & 182 & 168 \\
\hline
\emph{Goal} & 45 & 36  \\
\hline
\emph{Foul} & 3 & 10  \\
\hline
\emph{Penalty} & 3 & 1 \\
\hline
\end{tabular}
\quad
\begin{tabular}{|l|c|c|}
\hline
\textbf{Complex event}  & \textbf{Train Set}  & \textbf{Test set} \\
\hline
 \emph{Pass} &  2,670 & 2,389\\
\hline
 \emph{PassThenGoal} & 33 & 31 \\
\hline
\emph{FilteringPass} & 37 & 27 \\
\hline
 \emph{FilterPassThenGoal} & 4 & 4 \\
\hline
\emph{Cross} & 197 & 165 \\
\hline
\emph{CrossThenGoal} & 9 & 9 \\
\hline
 \emph{Tackle} & 1,413 & 1,130 \\
\hline
\emph{Shot} & 282 & 224 \\
\hline 
\emph{ShotThenGoal} & 41&36 \\
\hline
\emph{SavedShot} & 104 & 64 \\
\hline
\end{tabular}

\end{center}
\caption{Distribution of atomic and complex events (training and test set).  \label{table:list_of_events_dataset}}
\end{table*}


\section{Soccer event detection: a temporal logic approach}
 \label{sec:Eve}
 
The designed event detection system comprises two modules: an \textit{atomic event detector} and a \textit{complex event detector}. The first module takes as input the \textit{x} and \textit{y} coordinates of the players and the ball, and recognizes atomic (low-level) events through feature extraction and the application of predefined rules. The atomic events are stored in memory, and a temporal logic is then used to model and recognize low- and high-level complex events \cite{anicic2009event,etalis}.

The proposed system is capable of detecting overall five atomic events and 10 complex events, including all events defined in the ground truth except for fouls, penalties and goals, which would require additional information (such as the referee position and the $z$ coordinate of the ball). 

We adopt a methodology and notation similar to that used in \cite{khan2018soccer}, grounded on declarative logic, for the rule-based system. Briefly, an atomic event is defined as follows: 
	\[
	\begin{aligned}
	&SE=\langle ID, seType, t, \langle role_1, p_1 \rangle, ..., \langle role_i, p_i \rangle \rangle \\
	\end{aligned}
	\]
where ID is an event identifier, \textit{seType} is the type of the event, and \textit{t} is the time at which the event occurred; the event is associated to one or more objects, each identified as $p_i$ and associated to a specific $role_i$, \lia{which identifies the function played by the player in the event and is assigned automatically when the rule is verified}. The event can be associated to conditions to  be satisfied, e.g., based on the distance between the player and the ball.  

Complex events are built by aggregating other simple or complex events using temporal (temporal complex events) or logical operators (logical complex events): 
	\[
	\begin{aligned}
	&LCE=\langle ID, ceType, (t_s, t_e) ,L = \langle e_1 op e_2 op...op e_n \rangle\rangle \\
	&TCE= \langle ID, ceType,  (t_s, t_e) ,L = \langle e_1 THEN e_2 THEN...THEN e_n \rangle\rangle
	\end{aligned}
	\]

In all cases, \textit{ID} corresponds to the event identifier, \textit{ceType} to the event type,  $(t_s, t_e)$ is the time interval in which the event occurred, and $e_i$ is used to identify the sub-events. In the following, we do not differentiate between logical or temporal complex events. The main difference between our approach and that proposed in \cite{khan2018soccer} is that we model  time using intervals, rather than 
instants. \lia{Rule parameters were optimized using a genetic algorithm (see Section\ref{section:Evolutionary Algoritm}). }

\subsection{Atomic event detector}
\label{sec:atomic}

\subsubsection{Feature extraction}
Starting from the player and ball \textit{x} and \textit{y} positions,  the following features were calculated: \textit{velocity}, \textit{acceleration},  \textit{direction} with respect to the field, \textit{distance from the ball},   which players move, \textit{distance from the target line} of both teams, \textit{expected cross position on target line} and angle covered by the \textit{change of direction}. For a more detailed definition of the individual features, the reader is referred to the paper by Richly et al. \cite{richly2016recognizing}.


\subsubsection{Rules}

Atomic events are detected by applying a set of rules. Even if they are associated to a single time instant $t_{i}$, in order  to reduce the computational time and calculate stable values for the features, a sliding window approach was implemented: given a time instant $t_{i}$, the event $E_{i}$ is recognized if the corresponding rule is satisfied by the values in the interval $(t_{i},t_{i+k})$, where $k$ is equal to the window size. Feature extraction and rule checking were implemented in Python. Specifically, atomic events are defined as follows:
 
	\begin{enumerate}
		\item \textbf{KickingTheBall}  consists in a simple kick aimed at executing a cross, pass or shot. Starting from a position close to the player, the ball should move away from the player over the course of the window $k$, with a sudden acceleration and a final increased speed. 
		\[
		\centering
		\begin{aligned}
		&\langle ID, KickingTheBall, t,L=\langle\langle KickingPlayer, p_i\rangle ,\langle KickedObject,b\rangle \rangle \rangle \\
		&player(p_i), ball(b), Distance(p_i,b,t) <   T_{id_1} \\&
		\forall k = 1 \ldots n, D(p_i,b,t+k) < D(p_i,b,t+k+1),\\ &speed(b,t+n) <  T_{s_1}
		\exists  k | acceleration(b, t+k) <  T_{a_1}  \\
		\end{aligned}
		\]
		\item \textbf{BallPossession} is defined taking into account not only the player who has the control of the ball (i.e., the closest player), but also the player status (i.e., whether it is moving or not). Secondly, since the $z$ coordinate of the ball is not available, we used the ball speed to avoid accidentally triggering ball possession during cross events.
        \[
		\centering
		\begin{aligned}
		&\langle ID, BallPossession,t, L=\langle \langle PossessingPlayer, p_i\rangle ,\langle PossessedObject,b\rangle \rangle \rangle \\
		&player(p_i), ball(b), Distance(p_i,b,t) <  T_{id_2} \\ & \forall j \neq i, player (p_j),  D(p_j,b,t) >  D (pi,b, t)\\
		& \forall k = 1 \ldots n, D(p_i,b,t+k) < T_{id_2} \\
		& \forall k = 0 \ldots n, \forall j \neq i, team(p_j) \neq team(p_i), D(p_i, p_j,t+k) <  T_{od_2}, \\
		& speed(b, t+k) < T_{s_2} \\
		\end{aligned}
		\]
		
		\item \textbf{Tackle} occurs when a player (TacklingPlayer) tries to gain control of the ball against a player of the opposite team (PossessingPlayer). As a direct consequence, the presence of a member of the opposite team nearby is a condition to trigger the event.	
        \[
        \centering
        \begin{aligned}
        &\langle ID, Tackle,t, L= \langle \langle PossessingPlayer, p_i\rangle ,
        \langle TacklingPlayer, p_j\rangle \rangle ,\\
        &\langle PossessedObject,b\rangle \rangle, \\
        & player(p_i), player(p_j), ball(b),\\ 
        &Distance(p_i,b,t) < T_{id_3} \\
        & \forall u \ldots i, player (p_u), D(p_u,b,t) >  D (p_i,b, t) \\
        & \forall k = 1 \ldots n, D(p_i,b,t+k) <  T_{id_3} \\
        &\forall k = 0 \ldots n,
        \exists player(p_i) | D(p_i, p_j,t+k) <  T_{od_3}, team(p_i) \neq team(p_j), \\
        &speed(b, t+k) < T_{s_3} \\
        \end{aligned}
        \]

		\item \textbf{BallDeflection} occurs when the ball has a sudden change in direction, usually due to a player or the goalkeeper deflecting it. The ball in this event undergoes an intense deceleration reaching an area far from the deflecting player.
                  \[
		\begin{aligned}
		&\langle ID, BallDeflection,t,L= \langle
		\langle DeflectingPlayer, p_i\rangle 
		\rangle DeflectedObject,b\rangle \rangle  \rangle \\
		&player(p_i), ball(b), Distance(p_i,b,t) < T_{id_4} \\ & \forall k = 1 \ldots n,                D(p_i,b,t+k) < D(p_i,b,t+k+1), \\&speed(b,t+n) > T_{s_4}  \\ 
		& \exists k | acceleration(b, t+k) < -T_{a_4} \\
		\end{aligned}
		\]

		\item \textbf{BallOut} is triggered when the ball goes off the pitch.
      
      	\item \textbf{Goal} occurs when a player scores a goal.
		
	\end{enumerate}

	\subsection{Complex event detector}

    This module was implemented based on a temporal logic; specifically the Temporal Interval Logic with Compositional Operators (TILCO) \cite{TILCO} was used. TILCO belongs to the class of ITLs, where each event is associated to a time interval. TILCO was selected among  several available options because it implements both qualitative and quantitative ordering, and defines a metric over time: thus, we were able to impose constraints on the duration of the events, as well as to gather statistics on their duration. The ETALIS (Event TrAnsaction Logic Inference System) open source library, based on Prolog, was used for implementation \cite{etalis}. The complex event detector is characterized by few parameters, which were manually optimized on the training set.

    For the complex events, the rules were formalized as reported in the following.
        \begin{enumerate}
        \item \textbf{Pass} and \textbf{Cross} events occur when the the ball is passed between two players of the same team, and hence can be expressed as a sequence of two atomic events, KickingTheBall and BallPossession, where the passing and receiving players belong to the same team. A cross is a special case in which the ball is passed from the sideline area of the field to the goal area. An additional clause is added to the pass detection (not reported for brevity) to evaluate the position of the players, straightforward in our case as the coordinate system coincides with the field.      
        
        \[
    	\begin{aligned}
    	&\langle ID, Pass,(t,t+k),t, L= \langle 
    	ID, KickingTheBall,\\&  \langle KickingPlayer, p_i,t\rangle ,\langle KickedObject,b,t\rangle \rangle  \\
    	&THEN \langle ID, BallPossession, \langle PossessingPlayer, p_j,t+k\rangle ,  \\ & \langle PossessedObject,b,t\rangle \rangle \rangle \\
    	&player(p_i), player(p_j), ball(b), team(p_i) = team(p_j),  k < Th3  \\
    		\end{aligned}
    		\]
  
        \item \textbf{FilteringPass} allows to create goal opportunities when the opposite team have an organized defence. According to the UEFA definition, it consists of a pass over the defence line of the opposite team. In our definition, the player that receives the ball has to be, at the time the pass starts, nearer to the goal post than all the players from the opposite team. 
        \[
		\begin{aligned}
		&\langle ID, FilteringPass,t,t+k,t, L = 
		\langle ID, Pass,\langle PossessingPlayer,p_i,t\rangle ,\\&\langle ReceivingPlayer,p_j,t+k\rangle ,
		\langle PossessedObject, b, t\rangle \rangle \rangle  \\
		&player(p_i), player(p_j), ball(b), team(p_i) = team(p_j), \\
		&\forall k, player(p_k), team(p_k) \neq team(p_j), goal(g, p_k),\\& D(p_j, g, t + k) <  D(p_k, g, t + k) \\
		\end{aligned}
		\]

	    \item \textbf{PassThenGoal}, \textbf{CrossThenGoal}  and \textbf{FilteringPassThenGoal} are defined by the concatenation of two temporal sub-sequences: an alternation of Pass/FilteringPass/Cross followed by a Goal, where the receiver of the pass is the same player who scores.
       \item \textbf{Tackle}: as a complex event, it is a sequence of one or more atomic tackles, followed by a ball possession (which indicates the end of the action). A \textbf{WonTackle} terminates with the successful attempt to gain the ball by the opponent team. A \textbf{LostTackle} is obtained by the complementary rule.
        
        \item \textbf{ShotOut}, \textbf{ShotThenGoal} and \textbf{SavedShot} represent possible outcomes of an attempt to score. The SavedShot event, where the goalkeeper successfully intercepts the ball, is formalized as KickingTheBall followed by a BallDeflection or BallPossession, where the deflecting player is the goal keeper. 

	\end{enumerate}

\subsection{Event recognition from a multi-view camera setup}
\label{sec:archi}

	\begin{figure}[b!]
	\centering
		\includegraphics[width=\linewidth]{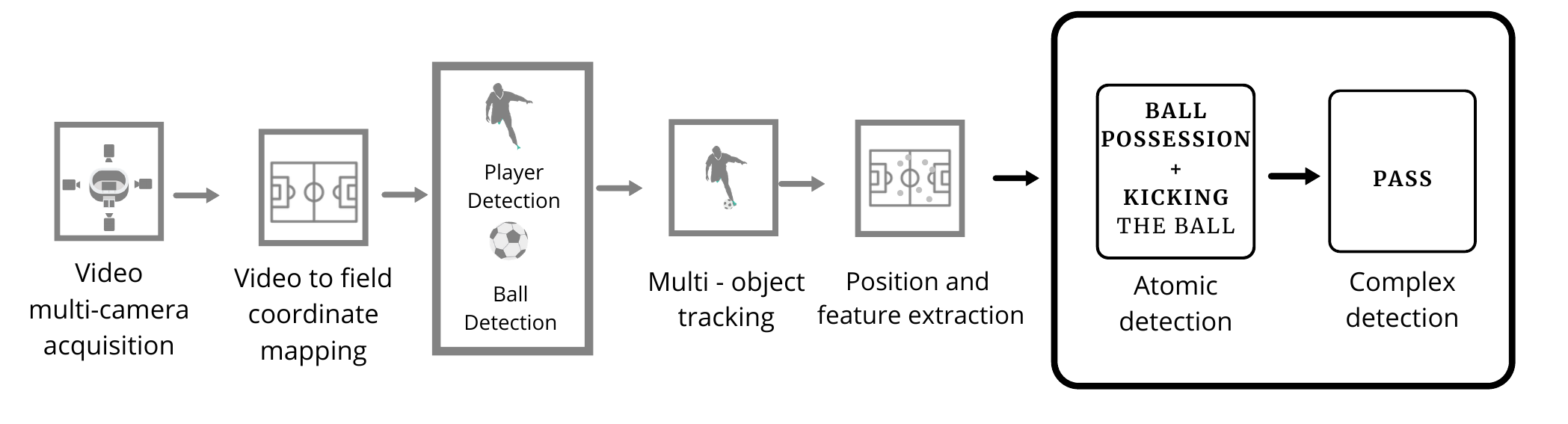}
		\caption{Deployment of the proposed system in a real-life scenario.}
		\label{Figure:flow_analysis}
	\end{figure}

\lia{
In a real setting, spatio-temporal data would need to be extracted from a multi-view video stream using a multi-object detection and tracking system (see Figure \ref{Figure:flow_analysis}). A multi-camera setup is required in order to solve occlusions and cover the entire playing field. For instance, Pettersen et al. used three wide-angle cameras to cover the Alfheim stadium \cite{Pettersen2014}; modern acquisition setup like Intel True View\textsuperscript{\textcopyright} include up to 38 5K cameras. The players and the ball can be detected using e.g., Single Shot Detector or another real-time object detector \cite{khan2018soccer,rematas2018soccer}. Pixel coordinates are then mapped to the field coordinate systems using a properly calibrated setup; alternatively, field lines can be used to estimate the calibration parameters \cite{rematas2018soccer}. For accurate event detection the system should be able to distinguish and track different players, assign them to the correct team, and minimize identity switches during tracking. For instance, certain events can only occur between players of the same team, other   between players of competing teams. Developing the detection and tracking system is beyond the scope of this paper. Instead, we exploit the game engine to log the position of the players and the ball at each frame, and focus on the final event detection step, which is further divided into atomic and complex event detection. }

\section{Experimental results}
\label{sec:results}

In this section, the evaluation protocol and the experimental results of the proposed detector on the SoccER dataset are reported. We focus first on the detection of atomic events, for which optimal parameters were found by means of a multi-objective genetic algorithm. Starting from the optimal solution of the atomic event detector, the performance of the complex event detector is analyzed and compared with the state of the art. 

\subsection{Evaluation protocol}

A ground truth atomic event is detected if an event of the same type is found within a temporal window of three frames. For complex events, we use the common OV20 criterion for temporal action recognition: a temporal window matches a ground truth action if they overlap, according to the Intersection over Union, by 20\% or more \cite{gaidon2011actom}. For each event, we calculate the recall, precision and F-score. 

\subsection{Parameter optimization: an evolutionary strategy}
\label{section:Evolutionary Algoritm}

Genetic or evolutionary algorithms are effective techniques for parameter optimization, as they only require the ability to evaluate the fitness function and are applicable when an analytic formulation of the loss is not available \cite{morra2018optimization}. In our case, the fitness value is based on the weighted average of the recall and precision metrics over all the event types. Since precision and recall are competing requirements, we opted for a multi-objective implementation, the Strength Pareto Evolutionary Algorithm or SPEA2 \cite{ZLTh_01_SPE}. SPEA2 is a Pareto-based optimization technique which seeks to approximate the Pareto-optimal set, i.e., the set of individuals that are not dominated by any others, while maximizing the diversity of the generated solutions. 

Each individual's genome encodes the set of 16 parameters associated to all rules. \francesco{The parameters of each rule are defined in Section \ref{sec:atomic} (i.e.,Inner Distance($T_{id_N}$), Outer Distance($T_{od_N}$), 
speed($T_{s_N}$) and accelleration($T_{a_N})$, where $N$ ranges from 1 to 4). In addition, the window for each rule is separately optimized.} Finally, since the rules are not mutually exclusive, the order in which they are evaluated is also encoded using the Lehmer notation. A range and discretization step is defined for each real-valued parameter to limit the search space. All window sizes are limited in the range 3--30 frames (with unitary step), all thresholds on speed were limited in the range 1--15 with step 1.0, and all thresholds on distance were limited in the range 0.1--2.0 \lia{meters} with step 0.1. 
The genetic algorithm was run for 50 generations starting from a population of 200 individuals; genetic operators were the BLX-0.5 crossover \cite{alcala2007multi}, with probability 90\%, and random mutation with probability 20\%. An archive of 100 individuals was used to store the Pareto front.  The optimal parameters were determined on the training set and evaluated on the testing set. The experiment was repeated twice to ensure, qualitatively, the reproducibility of the results. \lia{
Genetic algorithms are sensitive to random initialization and more runs would be needed to estimate the variability in the results. }

\begin{figure}[t]
\centering
\begin{tabular}{cc}
\subfloat[\label{Figure:Results of Archive generation number 50}]{\includegraphics[width=.45\textwidth]{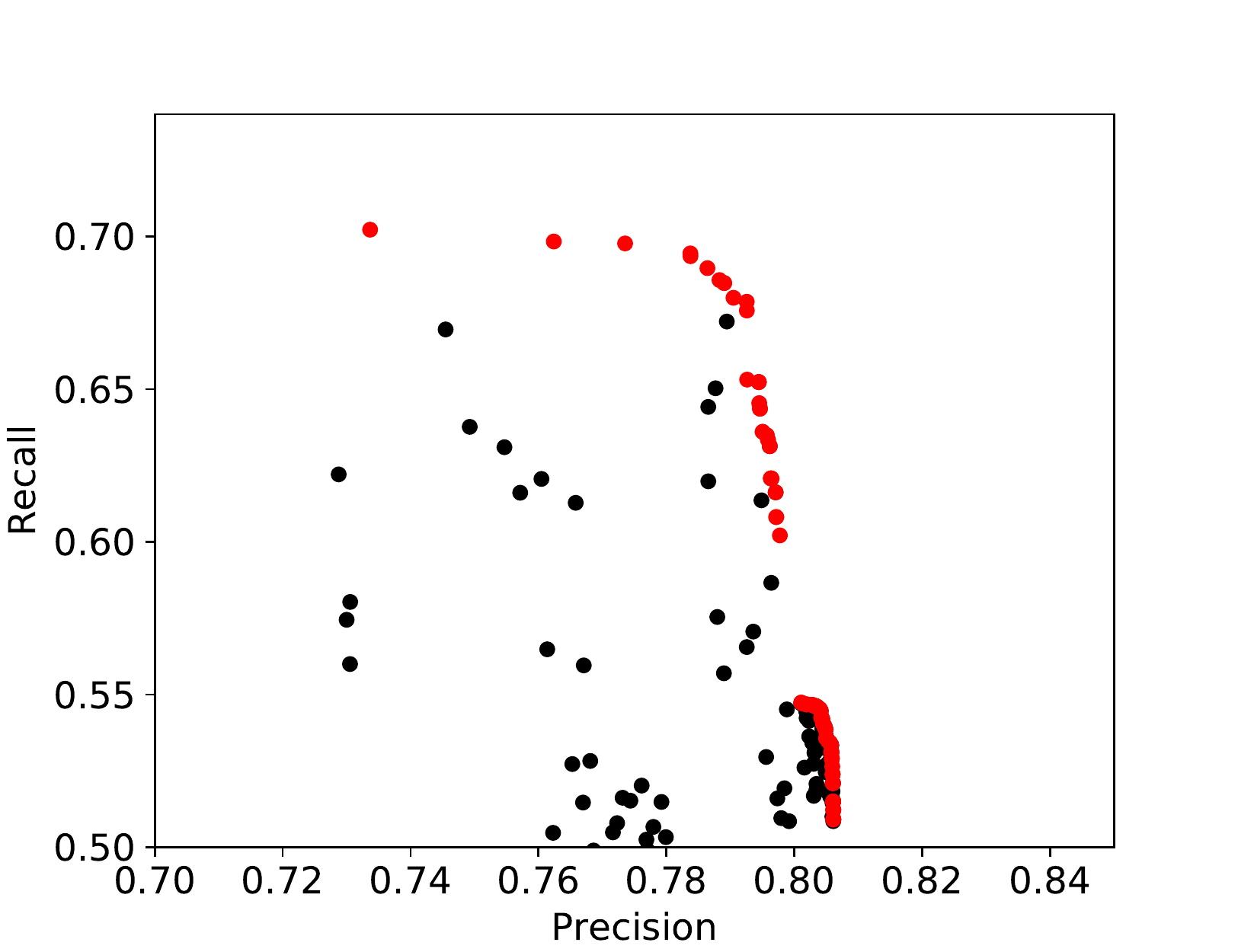}} &
\subfloat[\label{Figure:Complex event comparision}]{\includegraphics[width=.55\textwidth]{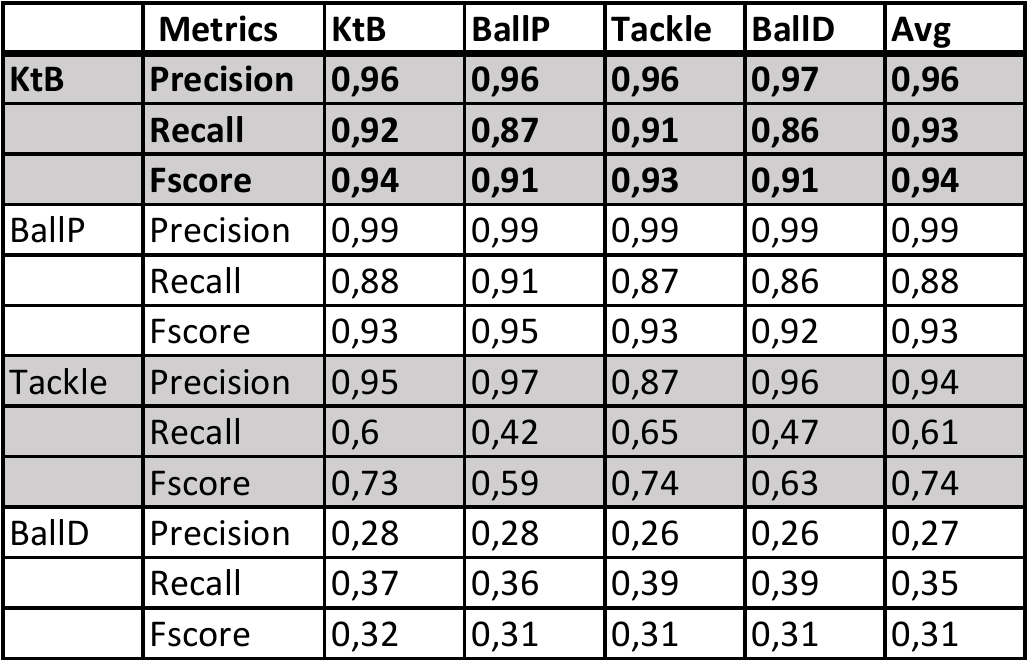}} 

\end{tabular}
\caption{Visualization of the Pareto front after 50 generations (a) and performance of the four best solutions generated (b). In (a) each dot represents a possible solution, and those belonging to the Pareto front are highlighted in red. In (b), each column represents the solution which maximize the F-score with respect to a specific event: KickingTheBall (KtB), BallPossession (BallP), Tackle and BallDeflection (BallD). For each event (row), the average performance is reported in the last column.}
\label{Figure:map_atomic_event}
\end{figure}

The final set of solutions, which approximate the Pareto front, is shown in Fig.~\ref{Figure:Results of Archive generation number 50}. The four solutions which maximize F-score for each event are compared in Fig. \ref{Figure:map_atomic_event}. The BallOut event (not reported) reaches perfect scores for all parameter choices. The easiest events to detect are KickingTheBall, with an average F-score of 0.94, and BallPossession, with an average F-score of 0.93. For Tackle, the average precision is high (0.94), but the recall is much lower (0.61). The worst result is obtained for BallDeflection, with values of F-score consistently lower than 0.4. Some events are more difficult to detect based on positional data alone, i.e., without considering the position of the joints or the action performed by the players \cite{pattern}. The best performing solution for the Tackle event (0.65 vs. 0.42 recall) corresponds to a lower recall for BallPossession (0.91 vs. 0.87), largely due to the similarity between the two classes; the difference in absolute values is easily explained by the higher frequency of BallPossession events.

\subsection{Parameters Evolution}
   The distribution of the parameter values at different iterations provides additional insight on the role of each parameter and the effectiveness of each rule. Two competing factors are responsible for the convergence towards specific parameter values: lack of diversity in the population, leading to premature convergence, and the existence of a narrow range of optimal values for a given parameter. We ruled out the first factor by repeating the experiment: we assume that parameters that converge to a stable value across multiple runs are more critical to the overall performance, especially if they are associated to high detection performance.

\begin{figure}[t]
\centering
\begin{tabular}{cccc}
\subfloat[Window size \label{Figure:WindowSize1}]{\includegraphics[width=.3\textwidth]{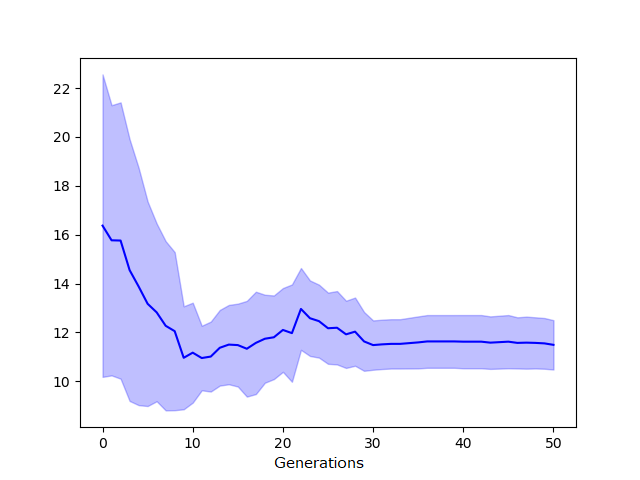}} &
\subfloat[Ball-player distance \label{Figure:InnerDistance1}]{\includegraphics[width=.3\textwidth]{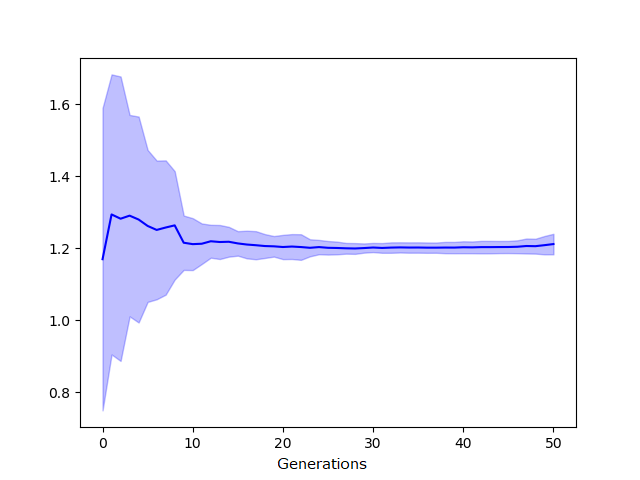}} &
\subfloat[Ball Speed \label{Figure:speed1}]{\includegraphics[width=.3\textwidth]{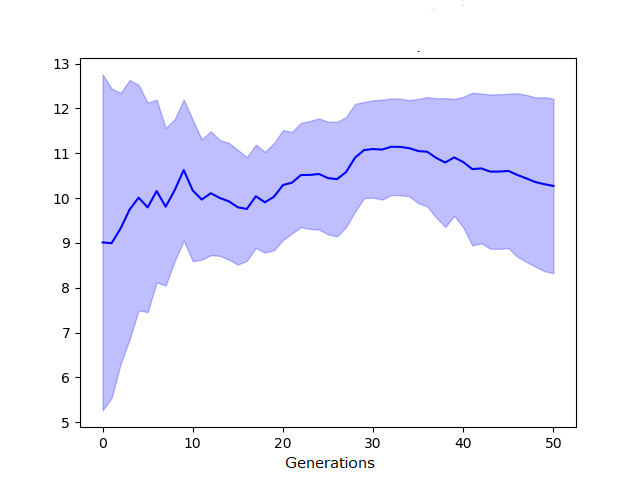}} &

\end{tabular}
\caption{Distribution (mean and standard deviation) of each parameter of the rule KickingTheBall  calculated over the entire population at each iteration.}
\label{fig:param_evoluion}
\end{figure}

   Let us consider for instance the parameters for the KickingTheBall rule, represented in Fig. \ref{fig:param_evoluion}. The window size and distance threshold both converge to a very narrow range, suggesting that a strong local minimum was found. On the other hand, the threshold on the ball speed appears less critical. 
   
   Other parameters tend to behave in a similar way, although there are exceptions. Generally speaking, the system is very sensitive to the distance thresholds, and in fact they converge to very narrow ranges for all events except BallDeflection (results are not reported for brevity). For most events, the window size has a larger variance then KickingTheBall and, in general, the rules  seem quite robust with respect to the choice of this parameter.
   
    The existence of an optimal parameter value is not necessarily associated to a high detection performance: for instance, the distribution of the acceleration threshold for the BallDeflection has a very low standard deviation and very high mean (not shown), as the change of direction usually causes an abrupt acceleration. At the same time, acceleration alone is probably not sufficient to recognize the event. Finally, the order in which the rules are processed does not seem to play a fundamental role.

\subsection{Overall performance}

\begin{figure}[t]
	\centering
		\includegraphics[width=\textwidth]{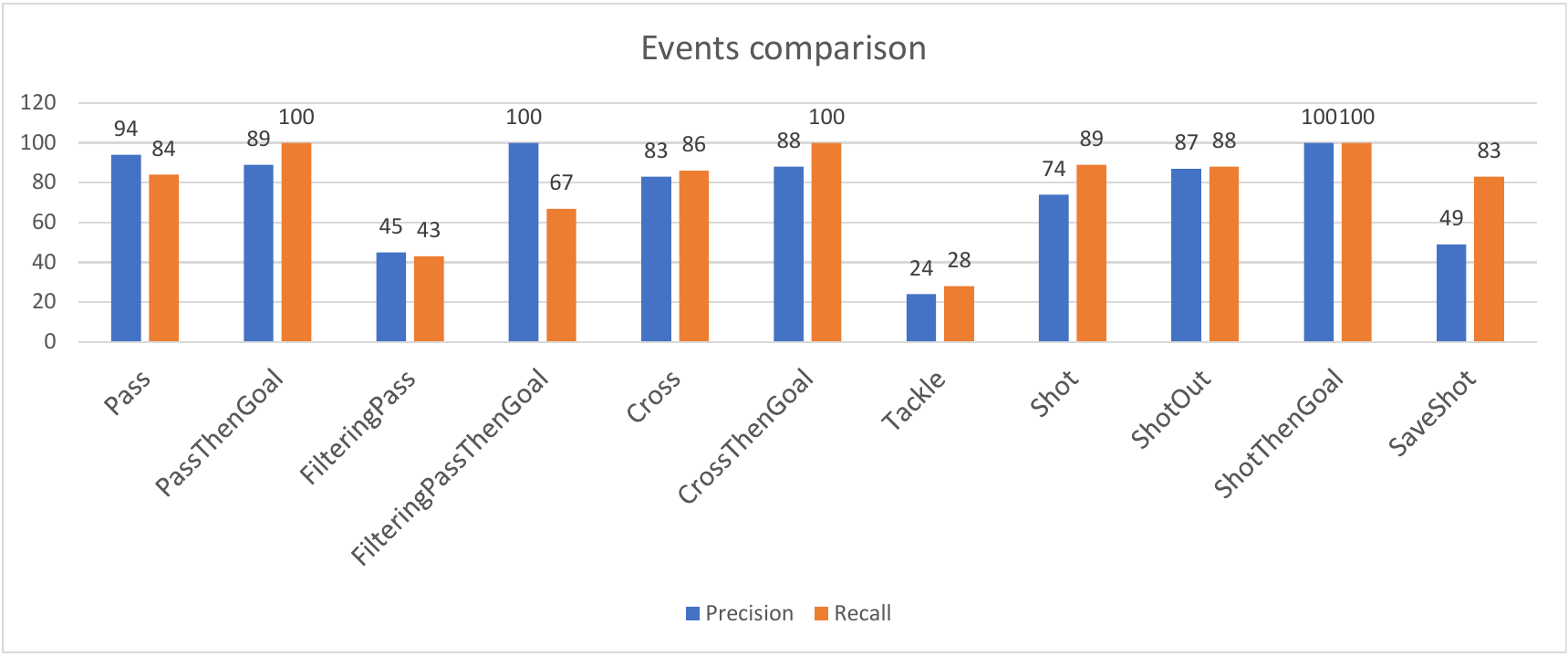}
		\caption{Precision and Recall for each complex event}
		\label{Figure:Comparison1}
	\end{figure}

The performance for complex events (precision and recall) is reported in Fig. \ref{Figure:Comparison1}. In eight out of 11 cases, the system was able to reach an F-score between 0.8 and 1. Sequences of events, such as passes that result in a goal, can be detected effectively. However, performance suffers when the detection of the atomic events is not accurate, e.g., for Tackle and SavedShot, which depend on the atomic events Tackle and BallDeflection, respectively.

Comparison with previous literature is difficult due to differences in the datasets, experimental settings, and types of events. Few previous works were based on positional data, extracted either from wearable trackers or using cameras covering the entire field \cite{richly2016recognizing,networks,pattern}. In the latter case, the accuracy of the positional data may further vary, depending on whether the ball and players are manually identified \cite{richly2016recognizing} or detected by a multi-object detector and tracker \cite{pattern}.

Despite these limitations, in Table \ref{Table:Final_Comparison} we attempt a comparison for two events: pass (complex event) and kicking the ball (atomic event). For both events, our results are comparable or better than previous literature, confirming that the proposed events can be successfully detected using (i) positional data (as in \cite{richly2016recognizing,networks}) and (ii) temporal logic (as in \cite{khan2018soccer}). It should be noticed that the SoccER dataset is much larger than those used in competing approaches, 
, including 1,203 passes and 1,728 kicking the ball events: datasets included in Table \ref{Table:Final_Comparison} range between 14 and 134 events).

	\begin{table}[tbh] 
\begin{center}
\begin{tabular}{p{2.2cm}|p{2.2cm}|p{3cm}|c|c|c}
\hline
\textbf{Solution} & \textbf{Input}  & \textbf{Method} & \textbf{Precision} & \textbf{Recall} & \textbf{F-score} \\
\hline
\multicolumn{6}{c}{\textit{kicking the ball} }  \\
\hline
Richly (2017) \cite{richly2016recognizing} & positional data & feature extraction + neural networks & 95\% & 92\% & 93\%\\
\hline
Khan (2018) \cite{khan2018soccer} & broadcast video & object detection + temporal logic & - & 92\% & 89\%  \\
\hline
Ours  & positional data & temporal logic & 96\% & 93\% & 94\%\\
\hline
\multicolumn{6}{c}{\textit{pass} }  \\
\hline
Khan (2018) \cite{khan2018soccer} & broadcast video & object detection + temporal logic & 94\% & 84\% & 89\%  \\
\hline
Richly (2016) \cite{richly2016recognizing} & positional data & feature extraction + SVM  & 42.6\% & 64.7\% & 51\%  \\
\hline
Lee (2017) \cite{pattern} & Fixed camera, entire pitch & Action recognition + finite state machine & - & 60\% & -  \\
\hline
Ours  & positional data & temporal logic &96\% & 93\% &  94\% \\
\hline

\end{tabular}
\caption { Comparison between state of the art and proposed approach. \label{Table:Final_Comparison}}
\end{center}
\end{table}

	\section{Discussion and conclusions}
\label{sec:discussion}
Event recognition in soccer is a challenging task due to the complexity of the game, the number of players and the subtle differences among different actions. In this work, we introduce the SoccER dataset, which is generated by an automatic system built upon the open source Gameplay Football engine. With this contribution, we strive to alleviate the lack of large scale datasets for training and validating event recognition systems. We modified the Gameplay Football engine to log positional data, as could be generated by a fixed multi-camera setup covering the whole field. Compared to the use of broadcast footage, we are thus able to consider the position of all players at once and model sequences of complex and related events that occur across the entire field. In the future, the game engine could be further extended to generate data on-the-fly, e.g., for the training of deep neural networks. 

A second contribution is the design and validation of ITLs for soccer event recognition. ITLs provide a compact and flexible representation for events, exploiting readily available domain knowledge, given that sports are governed by a well-defined set of rules. The capability of reasoning about events is key to detect with high accuracy complex chains of events, such as ``passes that resulted in a scored goal'', bypassing the need for extensive training and data collection. Relationships between events are also easy encoded.

\lia{Spatio-temporal positional data in the SoccER dataset may be more accurate than those extracted from real video streams, as explained in Section \ref{sec:archi}. Previous works reported a tracking accuracy of about 90\% for the players and 70\% for the ball in a multi-camera setup \cite{pattern}. It is possible to accurately and fairly compare different event detection techniques using synthetic data. Nonetheless, investigating the performance on real video streams, in the presence of noise, will require further investigation. }

In conclusion, we have shown that ITLs are capable of accurately detecting most events from positional data extracted from untrimmed soccer video streams. Future work will exploit the SoccER dataset for comparing  other event detection techniques, for instance based on machine learning \cite{giancola2018soccernet}.

%
%
%
\bibliographystyle{splncs04}
\bibliography{iciar_foot_bib}
\end{document}